\documentclass[10pt,journal,compsoc]{IEEEtran}

\ifCLASSOPTIONcompsoc
  \usepackage[nocompress]{cite}
\else
  \usepackage{cite}
\fi
\ifCLASSINFOpdf

\else
 \fi

\usepackage{graphicx}
\usepackage{xcolor}
\usepackage{url}

\begin{document}

\title{Cluster Activation Mapping with Applications to Medical Imaging}

\author{Sarah~Ryan, Nichole~Carlson, Harris~Butler, Tasha~Fingerlin, Lisa~Maier, Fuyong~Xing}

\IEEEtitleabstractindextext{%
\begin{abstract}
An open question in deep clustering is how to understand what in the image is creating the cluster assignments. This visual understanding is essential to be able to trust the results of an inherently complex algorithm like deep learning, especially when the derived cluster assignments may be used to inform decision-making or create new disease sub-types. In this work, we developed novel methodology to generate CLuster Activation Mapping (CLAM) which combines an unsupervised deep clustering framework with a modification of Score-CAM, an approach for discriminative localization in the supervised setting. We evaluated our approach using a simulation study based on computed tomography scans of the lung, and applied it to 3D CT scans from a sarcoidosis population to identify new clusters of sarcoidosis based purely on CT scan presentation.
\end{abstract}

}

\maketitle

\IEEEdisplaynontitleabstractindextext

%
\IEEEpeerreviewmaketitle

\IEEEraisesectionheading{\section{Introduction}\label{sec:introduction}}

With technological advancements and decreasing costs, vast amounts of medical imaging datasets are becoming available for clinical and scientific interpretation. However, annotation of imaging data is difficult due to its size and complexity, costly as it currently requires the time and expertise of a limited set of trained readers, and subjective as the annotations can vary based on the reader. As a result, large imaging datasets may be unlabelled; that is, there is no annotation of the scan of interest to make it recognizable in some way, such as normal or diseased, disease severity, or type of disease. In addition, annotations from visual assessment or scoring may not adequately capture the complex patterns observed in the data. In either case, objective, data-driven methods that can intelligibly and trustworthily assign labels to medical imaging datasets are needed. In this work, we extend an existing deep clustering framework \cite{xie2016unsupervised, guo2017deep, mrabah2019deep} by coupling it with our novel cluster activation mapping technique \cite{wang2019score} to learn the best clusters of scans from medical imaging datasets and understand the reason for a scan's cluster assignment by providing a visual representation of cluster importance. 

\begin{table}[h]
\centering
\caption{Radiomic feature extraction options to cluster medical imaging data. DC = deep clustering. CLAM = cluster activation mapping. }
\footnotesize
\label{tab:cluster}
\begin{tabular}{lccc}
  \hline
 & Traditional & Learned & Learned + CLAM \\ 
  \hline
  Objective  &  X & X & X \\
  Representative  &   & X & X \\
  Understandable  &  &  & X \\
   \hline
\end{tabular}
\end{table}

Given a large dataset of unlabelled 3D medical images represented by raw voxels, there are many objective, data-driven ways to define the feature space on which clustering is performed (Table \ref{tab:cluster}) \cite{xie2016unsupervised}.  A common method for feature extraction in medical imaging datasets is radiomic analysis \cite{ryan2019radiomic}, a quantitative imaging technique where large amounts of textural features are extracted from medical images. Traditional radiomic analyses rely on the extraction of hundreds of pre-specified quantitative features, which includes first-order histogram features, second-order grey-level co-occurrence matrices (GLCM), grey-level run-length matrices (GLRLM), fractal analyses, and others \cite{kolossvary2018cardiac}. This collection of radiomic features may not be the best representation of underlying disease abnormalities in medical imaging data, as these features are pre-specified and common across all disease types. In contrast, \textit{learned} radiomics is an emerging subfield within radiomics \cite{afshar2018hand} that relies on the extraction of features which are learned from an autoencoder, a type of unsupervised machine learning architecture. The autoencoder learns the best representation of the data using a set of features, known as deep embeddings or latent features \cite{xie2016unsupervised}, which are in a reduced dimension compared to the size of the original image. Thus, learned radiomic features can be advantageous and more representative over traditional radiomic features for datasets with diseases that have not been classified by a domain-specific expert, or are too complex to be visually classified. 

Clustering methods using learned radiomic features are referred to as deep clustering \cite{mrabah2019deep}, and have found success in many applications. However, an open question in deep clustering is how to understand what in the image is creating the cluster assignments. This visual understanding is essential to be able to trust the results of an inherently complex algorithm like deep learning. This is especially important when the cluster assignments may be used to inform decision-making or create new disease sub-types, and clinicians want to better understand the clustering results.

\begin{figure}
    \centering
    \includegraphics[width=0.5\textwidth]{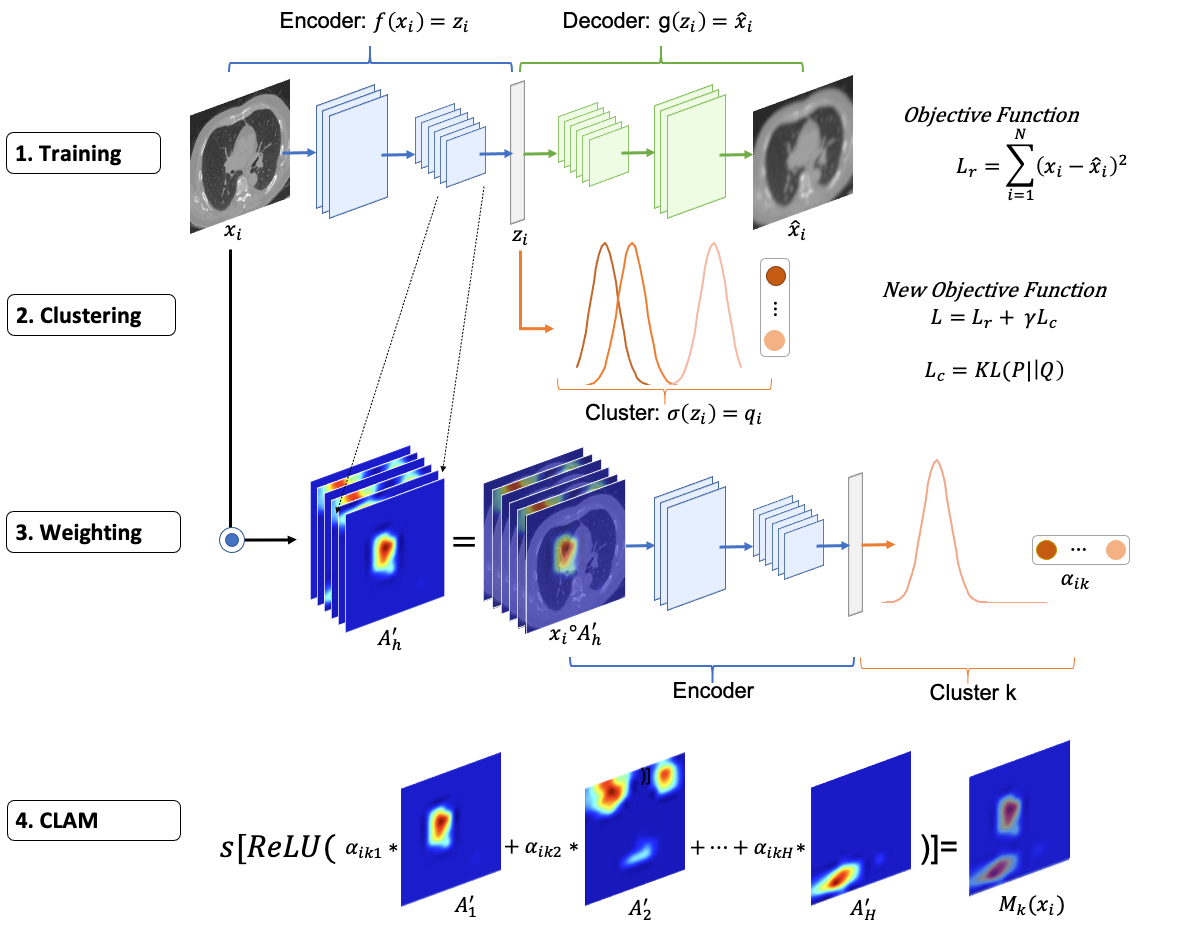}
    \caption{Steps used to create a cluster activation map (CLAM) for a particular image, which highlights the discriminative regions on the original image to identify why the it was assigned to a particular cluster.}
    \label{fig:arch}
\end{figure}

In the supervised setting, approaches have been developed to explain the predictions of machine learning models. A popular approach is class activation mapping (CAM). CAM uses a linear combination of the activation channels, or maps, from a particular convolutional layer to identify the discriminative regions from the original image used to predict a particular category \cite{zhou2016learning}. The original CAM uses a global average pooling (GAP) layer prior to the softmax to obtain the weights of each of the activation channels. However, the incorporation of a GAP layer may result in a decrease in performance. Thus, other methods, which do not require the GAP layer, have been developed \cite{selvaraju2017grad, wang2019score}. One such method, Score-CAM, calculates the weight of each activation channel using a global contribution of the corresponding input features \cite{wang2019score}. Score-CAM has been shown to achieve better performance for discriminative localization for classification than existing methods \cite{wang2019score}. However, none of the CAM methods have been applied in the unsupervised deep clustering framework. In this work, we develop novel methodology to generate CLuster Activation Mapping (CLAM) which combines an unsupervised deep clustering framework with a modification of Score-CAM, an approach for discriminative localization in the supervised setting.

The remainder of this paper is organized as follows. In Section \ref{deep}, we review deep clustering methodology, followed by the introduction of our CLAM methodology in Section \ref{clam}. The performance of the method is evaluated by performing a simulation study based on lung images in Section \ref{lungsims}. In Section \ref{camapp}, we apply our methodology to the CT scans from a population of patients with pulmonary sarcoidosis to obtain new subtypes based on abnormal CT presentations. We conclude with a discussion in Section \ref{camconc}. 

\section{Related Work}
\label{deep}

To understand CLAM, we first review deep clustering methodology in which the feature learning and clustering are performed jointly. In these techniques, an autoencoder is trained to learn representative features from the data. Then, a clustering layer is added to identify cluster assignments based on learned features. These phases of deep clustering are described in detail below. 

\subsection{Phase 1: Training}

An autoencoder is a type of unsupervised neural network for image reconstruction, whose outputs are the reconstructed inputs. It consists of two parts, an encoder and decoder. The encoder is a function that maps a set of images from their original space into a latent space, and the decoder is a function that maps the set of features from the latent space back into their original image space. In notation, assume we have sets of images $X=\{{x_i} \in R^S\}_{i=1}^{N}$, features $Z=\{z_i \in R^J\}_{i=1}^{N}$ and reconstructed images $X' = \{x'_i \in R^S\}$, where $x_i$ is the $i^{th}$ image, $N$ is the number of images, $S$ is the number of pixels (or voxels in 3D) per image, and $z_i$ is a $J$-dimensional set of latent features for the $i^{th}$ image. Then, the encoder and decoder are represented as follows: 

\begin{equation}
\begin{array}{l}
{f_{\theta_e}: X \to Z}\\
{g_{\theta_d}: Z \to X'}
\end{array}
\end{equation}

\noindent where $\theta_e$ and $\theta_d$ are the learnable parameters for the encoder and decoder, respectively, and for a given image, its reconstructed images is given by: $x'_i = g[f(x_i)]$. When the number of features in the latent space is less than the number of pixels in the image space ($J<S$), the autoencoder can be used as a type of dimension reduction. To optimize these parameters, the autoencoder is trained to minimize a reconstruction error, also known as a loss function. In the pre-training phase, the mean squared error is used as the reconstruction loss $L_r$: 

\begin{equation}
 L_r=\sum_{i=1}^N \sum_{s=1}^S \left[x_i(s) - x'_i(s)\right]^2
\end{equation}

For imaging data, we use convolutional layers within the autoencoders due to their ability to localize objects within the images \cite{zhou2016learning}, which are known specifically as convolutional autoencoders (CAE). 

\subsection{Phase 2: Clustering}
After training, we assign images to clusters during the clustering phase. To do this, we add a clustering layer onto the encoder, and define a new function for mapping the set of images $X$ to the set of soft probabilities for cluster assignment $Q = \{q_i \in R^K\}$: 

\begin{equation}
\label{eq:sigma}
{\sigma_{\theta_c}: X \to Z \to Q}
\end{equation}

\noindent where $\theta_c$ are the learnable parameters, and $K$ is the number of clusters. We find the similarity between embedded point $z_i = f(x_i)$ and the latent cluster with centroid $\mu_k$, $k \in {1, \dots, K}$, using the student's t-distribution \cite{maaten2008visualizing} with one degree of freedom and normalized over all clusters $K$: 

\begin{equation}
\label{eq:qik}
q_{ik}=\frac{\left[1+\sum_{j=1}^J (z_{i,j} - \mu_{k,j})^2\right]^{-1}}{\sum_{k=1}^K\left[1+\sum_{j=1}^J (z_{i,j} - \mu_{k,j})^2\right]^{-1}}
\end{equation}

\noindent where $q_{ik}$ can be described as the probability of assigning image $x_i$ to cluster $k$. The cluster centroids in latent space $\mu_k $ are first initialized using k-means, then updated during training. For this clustering phase, the overall loss $L$ is modified by the addition of a clustering loss $L_c$ to obtain embedded features which are also discriminatory: 

\begin{equation}
 L=L_r + \gamma L_c
 \end{equation}

\noindent where $\gamma>0$ controls the contribution of the cluster loss, $L_c$ and is typically set at $\gamma = 0.1$ \cite{xie2016unsupervised}.  Here it is important to also keep the reconstruction loss $L_r$ to enable good representations of the images \cite{mrabah2019deep}. The cluster loss is often defined using the Kullback-Leibler (KL) divergence \cite{guo2017deep}: 

\begin{equation}
L_c=KL(P||Q)=\sum_{i=1}^N \sum_{k=1}^K p_{ik} log \frac{p_{ik}}{q_{ik}}
\end{equation}

\noindent where $P$ is a target distribution of clusters, and $KL(P||Q)$ is interpreted as the amount of information lost when Q is used to approximate P. The clustering loss aims to minimize the $KL$ divergence so that our estimated cluster distribution converges to the target cluster distribution. We follow \cite{xie2016unsupervised} to define the target distribution as: 

\begin{equation}
p_{ik}=\frac{q_{ik}^2/\sum_{i=1}^N q_{ik}}{\sum_{k=1}^K (q_{ik}^2/\sum_{i=1}^N q_{ik})}
\end{equation}

This target distribution is used as it strengthens soft cluster probabilities for each image (i.e. encourages high probabilities to become higher, and low probabilities to become lower) by putting more weight on the data points assigned to clusters with high-probability, as implemented in the equation by squaring the probability, $q^2_{ik}$, then normalizing by the sum of probabilities for cluster $k$ across all images. Additionally, it prevents large clusters from monopolizing the feature space by normalizing across the clusters, as implemented with the denominator \cite{xie2016unsupervised}. 

Alternative methods have been used for the clustering loss. Of note is a recent method by \cite{mrabah2019deep}, which is based on the mean squared error of the latent space with the cluster centroids as the target. This approach avoids the influence of low probability data points by defining conflicted and unconflicted sets, which consists of the data points assigned to clusters with low and high-probability, respectively; however, this adds an additional parameter to optimize, which can be difficult to train in the unsupervised setting. Furthermore, the increase in performance with existing datasets is negligible \cite{mrabah2019deep}. For these reasons, and due to excellent performance in our simulation study, we use the KL divergence, as defined above. 

In this clustering phase, training ends when the loss has converged, as some have defined by the maximum percentage of images whose cluster assignments go unchanged between iterations \cite{mrabah2019deep} or after a set number of iterations \cite{xie2016unsupervised}. In this work, we plot the loss across the iterations, observing when the loss has stabilized, as we found under simulation that some cluster assignments will go unchanged in consecutive iterations early on in the training despite the loss still decreasing at a steady rate. Once training ends, each image $x_i$ is assigned to the cluster with the highest probability of cluster assignment, that is, $k' = argmax_k(q_{ik})$. 

\subsection{Selecting the Optimal Number of Clusters}

Under this framework, the number of clusters, K, needs to be set. Most deep clustering methods assume $K$ is known \cite{xie2016unsupervised, guo2017deep, mrabah2019deep}. However, in a purely unsupervised approach, the number of clusters is not known. Various methods outside of the deep clustering literature exist to estimate $K$, including the average silhouette method \cite{rousseeuw1987silhouettes}, which we evaluate in our simulation study, among others \cite{tibshirani2001estimating, liu2010understanding, patil2019estimating}. Furthermore, in the deep clustering framework, there are two times at which the optimal number of clusters could be estimated: (1) after the pre-training but before the clustering phase, or (2) after the clustering phase. We explore the performance of estimating $K$ at both of the times in our simulation section.

\section{Methodology}
\label{clam}

\subsection{CLAM for a Single Image}
Once images are assigned to clusters, we desire to identify discriminative regions from each image to understand the components in the image that lead to its cluster assignment. This understanding is especially important when the cluster assignments may be used to inform decision-making or create new disease sub-types, and clinicians want to better understand the clustering results. Various methods have been developed to identify discriminative localizations on images for \textit{classification} \cite{zhou2016learning, selvaraju2017grad, wang2019score}. In this work, we modify Score-CAM \cite{wang2019score}, due to its success for classification, to identify discriminative regions specifically for \textit{clustering}.  

For either classification or clustering, a linear combination of the activation channels from a selected convolutional layer in a neural network are needed to create a CAM or CLAM, respectively. To modify CAM to work for clustering, we make changes to our neural network and the weighting algorithm of the activation channels (Figure \ref{fig:arch}). In our deep clustering framework, we used a CAE, which consists of an encoder and decoder. However, to frame our clustering architecture to imitate a convolutional neural network for classification which takes in an image and outputs a scalar, we remove the decoder and add on a clustering layer as in Equation \ref{eq:sigma}. Then, any of the convolutional layers in the encoder whose weights have been trained after the clustering layer is added will be useful to identify regions of the image which are discriminative for cluster assignment, and as such, can be used to create the CLAMs. As noted for CAM \cite{selvaraju2017grad}, the selection of a particular convolutional layer is a trade-off between the spatial resolution, which is typically higher in earlier layers, and discriminative ability, which is improved in later layers. In this work, we use the last convolutional layer in the encoder just prior to the clustering layer unless otherwise noted.

For the weighting algorithm of the activation channels, Score-CAM is based on the Channel-wise Increase of Confidence (CIC) measure \cite{wang2019score}, which, for a particular image, is the increase in the target score of the original image that has been masked by a particular channel, as compared to the target score of the baseline unmasked image. Here, the idea is that the target score is improved when certain areas are masked out of the original image \cite{chattopadhay2018grad}. However, we have found that our target score, which is the probability of cluster assignment fixed between $[0,1]$ (Equation \ref{eq:qik}), rarely improves when certain areas are masked out. Furthermore, if an image $i$ is assigned to a particular cluster $k$ with probability $p_{ik}$, the maximum increase in probability will be $1-p_{ik}$; for high $p_{ik}$, the maximum possible increase will be small, which may underestimate the confidence, or importance, of a particular channel. For these reasons, we remove the comparison to the baseline image and define a channel-wise confidence (CC) to weight the activation channels for creating CLAMs, rather than the CIC for Score-CAM. Formally, we have the following definition for CLAM: 

\textbf{Definition}. For a given CAE model with a clustering layer as described in Section \ref{deep}, we define a function $q_{ik} = \sigma_k(x_i)$ which takes in image $x_i$ and outputs scalar $q_{ik}$, the probability of assignment to cluster $k$ using the student's t-distribution as defined in Equation \ref{eq:qik}. Let $A$ be the activations from the last convolutional layer of $\sigma_k$, and $H$ be the number of channels in A. For a single image $x_i$ and channel $h \in \{1 \dots H\}$, the CC for cluster k, denoted $\alpha_{ikh}$, is calculated as: 

\begin{equation}
\alpha_{ikh}=\sigma_k\left(x_i \circ A'_{h}\right)\\
\end{equation}

\noindent where $\circ$ denotes the Hadamard product and $A'_{h}$ is the $h^{th}$ activation channel that has been upsampled to the same spatial resolution as the original image and min-max normalized to [0,1], creating a non-binary smooth mask for $x_i$. The CC can be interpreted as the probability of an image's assignment to cluster $k$ after the original image is masked by the upsampled activation channel. Then, the CLAM $M_k(x_i)$ for a single image $x_i$ and cluster $k$ is a linear combination of the activation channels, followed by a rectified linear unit (ReLU) and min-max normalization function: 

\begin{equation}
M_{k}(x_i) = s\left[ReLU\left(\sum_{h=1}^H \alpha_{ikh} A_h\right)\right]
\end{equation} 

Here ReLU is used as the activation function to remove features with negative weights, since we are only interested in observing which features have a positive influence on cluster assignment. Further, the min-max normalization scales the CLAMs to $[0,1]$, so that the pixel with the most influence for cluster assignment for each image is valued at 1 and the non- (or negatively) influential pixels are valued at 0. This normalization, also implemented in CAM and its variants (\url{https://github.com/haofanwang/Score-CAM}), removes the effects of gross influences related to the linear combination of activation channels, allowing for better comparisons of discriminative regions between images. We refer to this scaled-influence unit of measurement for the CLAMs as \textit{importance} throughout this paper for simplicity. Furthermore, although CLAMs can be created for all clusters for each image, we only create the CLAMs using the cluster $k$ with the highest probability of cluster assignment for each image.  

\subsection{CLAM for Multiple Images}
Although CAM was originally developed to highlight discriminative regions of single images for classification, we summarize CLAMs from groups of images, creating group-averaged CLAMs to highlight common regions across the images which are important for cluster assignment and providing an overall representation of the scans. Given the registration of medical images to the same coordinate space in most medical imaging analyses \cite{ryan2019template}, the group-averaged CLAMs are informative for detecting common discriminative regions across medical imaging datasets, and result in a better understanding of the image characteristics assigned to each cluster. The cluster-averaged CLAM, $\tilde{M}_k$, and population-averaged CLAM, $\tilde{M}$, are given by: 

\begin{equation}
\tilde{M}_k=\sum_{x_i \in C_k} M_{k}(x_i)
\end{equation}

\begin{equation}
\tilde{M}=\frac{1}{K}\sum_{k=1}^K\sum_{x_i \in C_k} M_{k}(x_i)
\end{equation}

\noindent where $C_k = \{x_i: k = k'_i\}$ and $k'_i=argmax_k(q_{ik})$.

\section{Experiments}
\label{lungsims}

\begin{figure}
    \centering
    \includegraphics[width=0.4\textwidth]{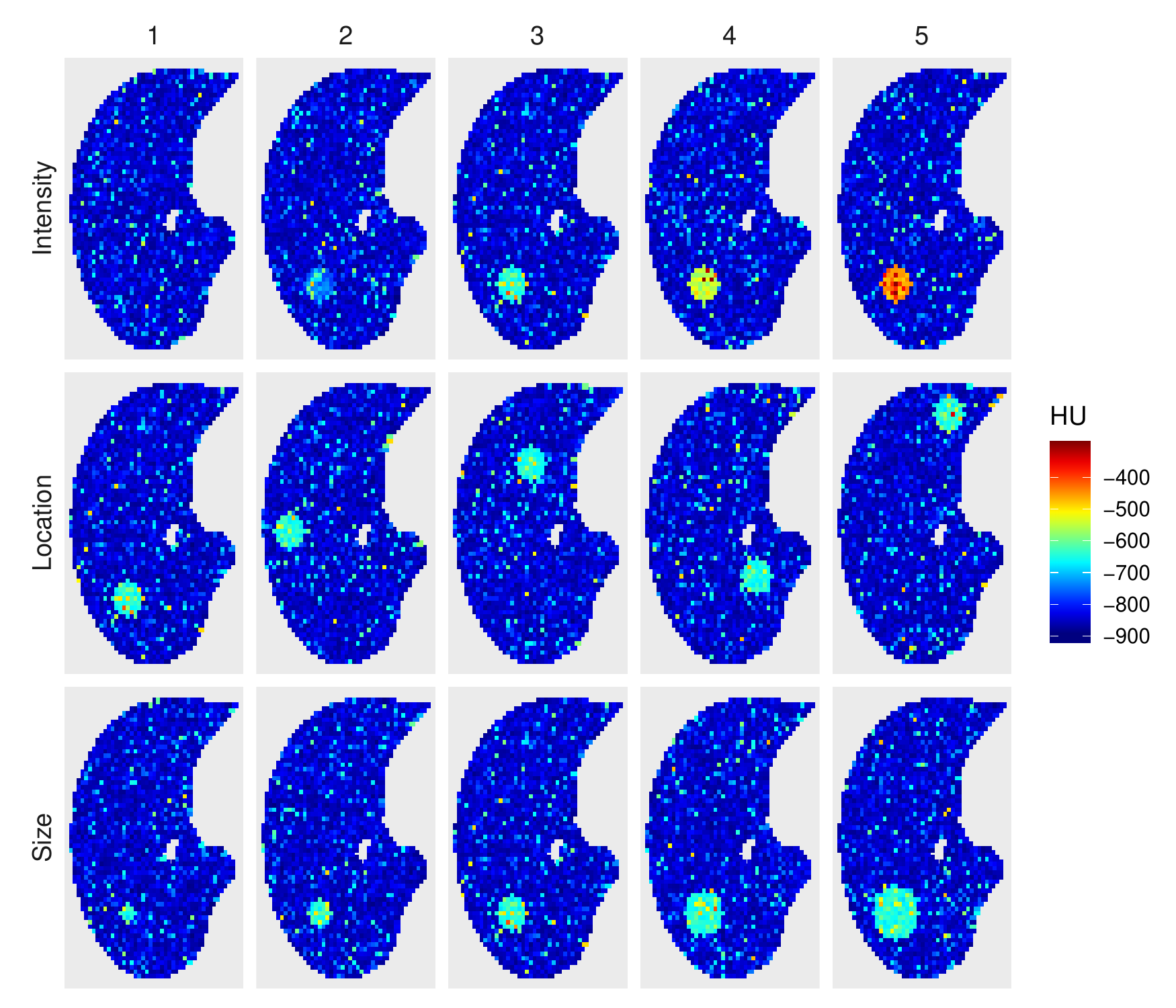}
    \caption{Example simulated lung slices, varying either the intensity, location or size of the simulated abnormality. }
    \label{fig:design_sim}
\end{figure}

\subsection{Design}

We examined the performance of our approach by conducting a simulation study. We simulated data under three different case scenarios, either varying the intensity, location or size to represent differing CT abnormality patterns (Figure \ref{fig:design_sim}). Across all scenarios, data was generated based on the median axial slice from the standard lung template that has been resampled to 3mm$^2$ voxel spacing, for ease of computation and visualization. Further, null background values were added to make the width and height of the image equal, which helps to remove isotropic tendencies from the CLAM, resulting in a 64x64 pixel image. The intensities of the CT abnormalities were simulated to range from 0 to 400 HU, in 100 unit increments; the locations varied across the lung slice as can be visualized in the middle row of Figure \ref{fig:design_sim}; the size varied from 2 to 6 pixel radii in increments of 1 pixel. In the scenarios, only one of characteristics (i.e. intensity, size or location) varied at a time, while the other characteristics were held constant. Random noise was added to the images from the empirical distribution of the standard healthy lung. Across scenarios, N=300 lung images were simulated such that the true number of clusters was set at $K=3, 4, 5$, resulting in a different number of images assigned to each cluster depending on $K$. 

A CAE with a clustering layer was fit to the data. The encoder network structure was $conv^2_{32} \to conv^1_{64} \to conv^2_{128} \to FC_{60}$ where $conv^h_n$ denotes a convolutional layer with $n$ filters, a stride length of h, and a kernel size of 4 x 4 as default. The decoder was a mirror image of the encoder. The autoencoder was trained end-to-end for 200 epochs using Adam \cite{kingma2014adam} and the leaky ReLU activation function. After training, the clustering layer described above was attached to the embedded layer of the encoder, and trained for 500 additional epochs. The number of clusters in the clustering layer was set at various $K = 2, \dots, 8$. After training, CLAMs were used to visualize discriminative regions for all images. 

\subsection{Results}

\begin{figure}
    \centering
    \includegraphics[width=0.4\textwidth]{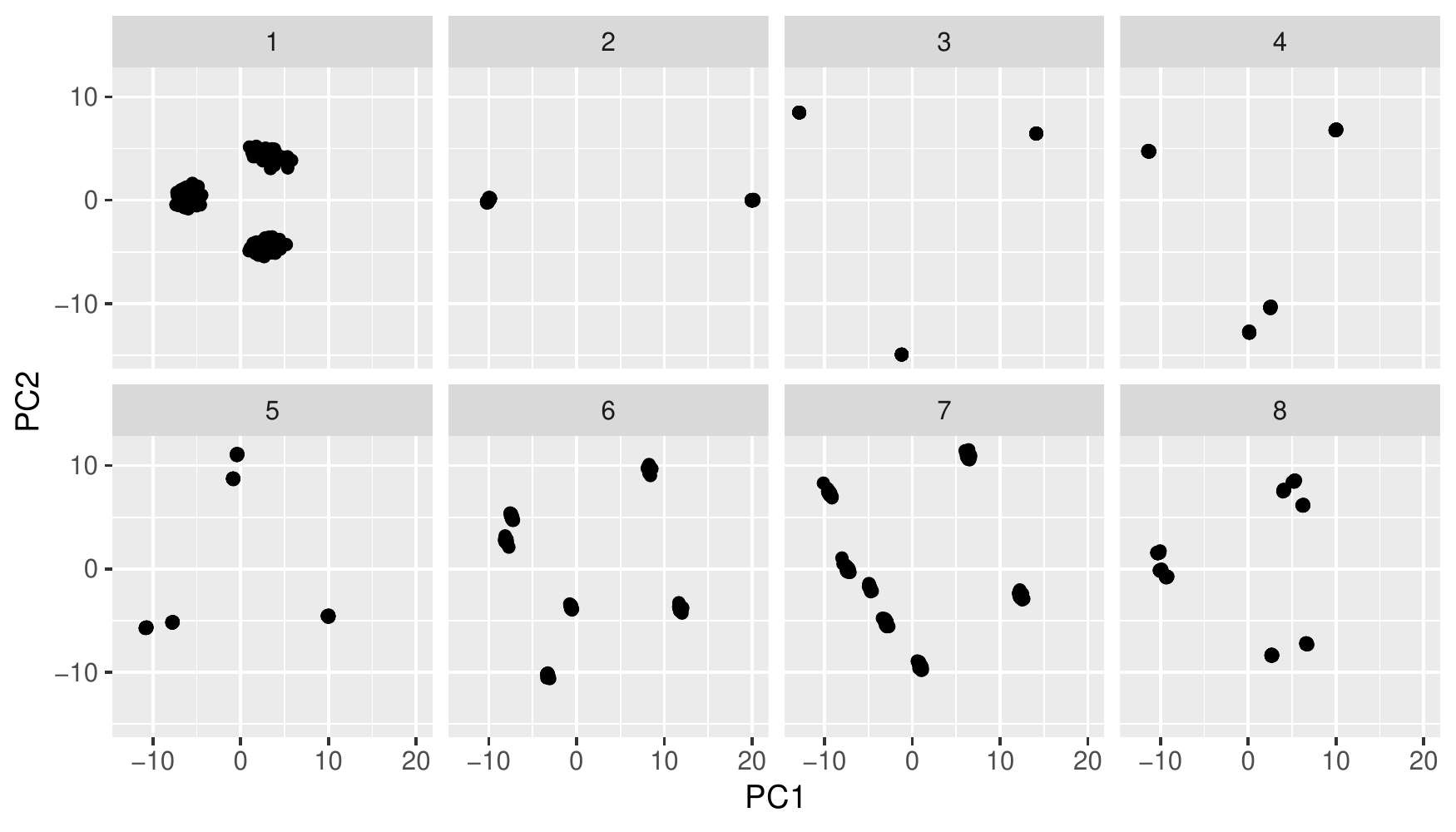}
    \caption{Diagram of features when true number of clusters $K=3$. The first subplot represents the feature space prior to clustering. Subplots $2-8$ represent the feature space after clustering, when set $K= {2,\dots, 8}$.} 
    \label{fig:pcs_sims}
\end{figure}

After the training phase, but before the clustering phase, the embedded features contain the most information from the images in a reduced dimension. In the top right subplot in Figure \ref{fig:pcs_sims}, the feature space after training is plotted in two-dimensions using principal components analysis for the scenario when the location of abnormalities on the image was varied and the true number of clusters was set at 3; in this plot, three clusters are observed. However, in the clustering phase, the latent space was influenced by the set number of clusters. Thus, the latent features, still containing the most information in a reduced dimension, formed into the number of clusters that was set, regardless of the true number of clusters, as is observed in Figure \ref{fig:pcs_sims}) when $K\neq3$. 

Across all scenarios, when $K$ was set to the true number of clusters, our method predicted the correct cluster assignment 100\% of the time. When fitting a larger number of clusters to the data than simulated, especially when varying the size or intensity of the abnormalities, the images switched clusters often. Interestingly, in these scenarios, the accuracies reached 100\% due to no images being assigned to the extra clusters. In contrast, all images may be assigned to one cluster, resulting in accuracies based on chance. This only occurred in our simulations when the fitted $K$ was not set at the true value, such as in the intensity scenarios when the true $K=4$ but was set at 2 or 5.

Using the silhouette method before the clustering phase, the true number of clusters was estimated as the optimal in seven of the nine scenarios, failing in the scenarios when the size of the abnormality varied and the true $K=4,5$, which were inaccurately estimated at $K=2$. After the clustering phase, only three of the nine scenarios estimated the true number of clusters as the optimal, for the size and location scenarios when true $K=3$ and for the intensity scenario when true $K=4$; none of the scenarios estimated the correct number of clusters post-clustering when the true $K=5$. The clustering phase resulted in features that formed the number of clusters that was set, regardless of true number of clusters. For these reasons, if the optimal number of clusters is not known \textit{a priori}, these results suggest that setting $K$ before the clustering phase may work better, although more research is needed. 

\begin{figure}
    \centering
    \includegraphics[width=0.2\textwidth]{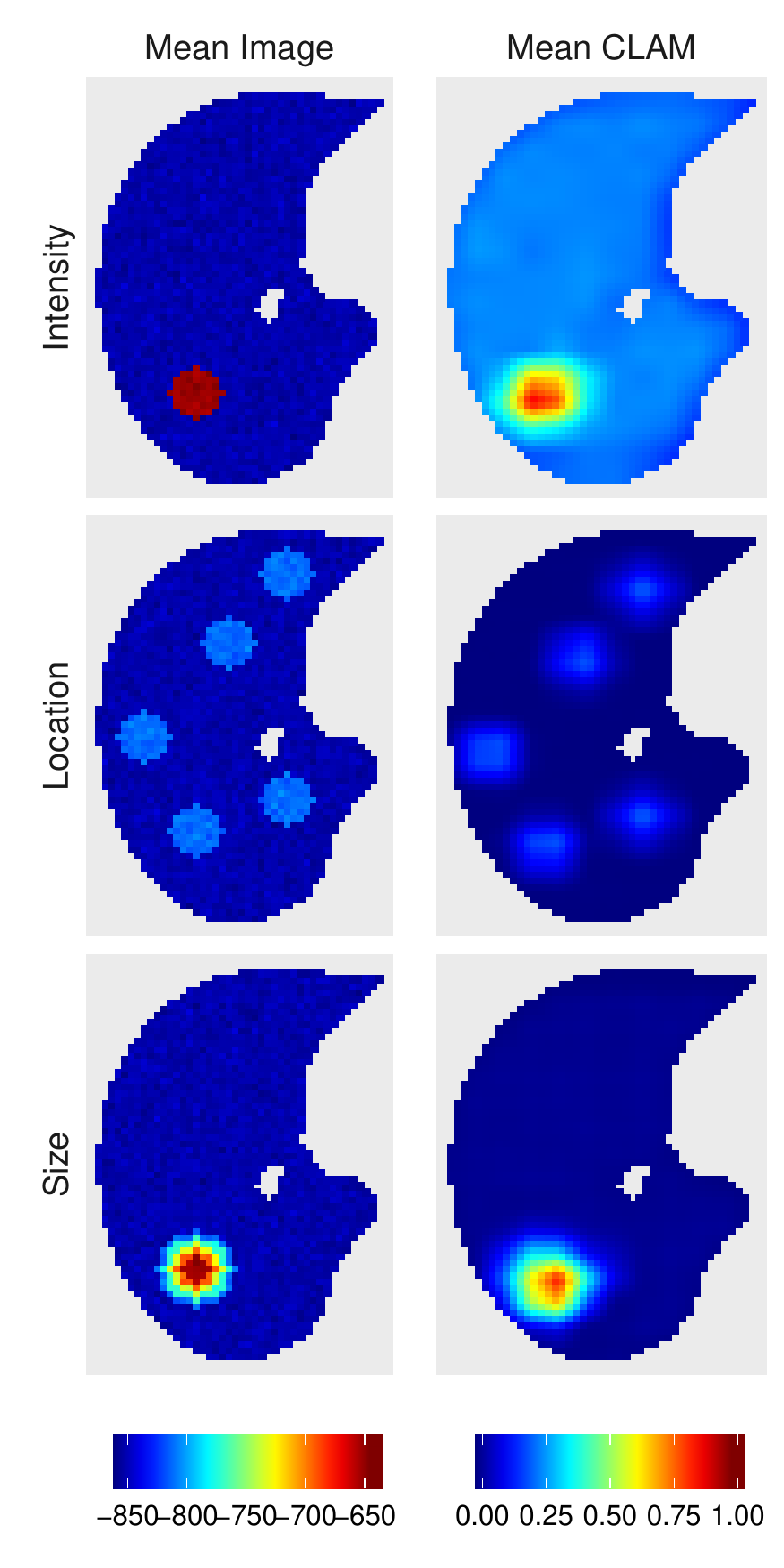}\\
    \caption{Population-averaged image (measured in HU) and CLAM (measured in units of importance) for each simulation scenario when true $K=5$ and the intensity, location, or size varies across the images. }
    \label{fig:ave_sim} 
\end{figure}

\begin{figure}
    \centering
    \includegraphics[width=0.4\textwidth]{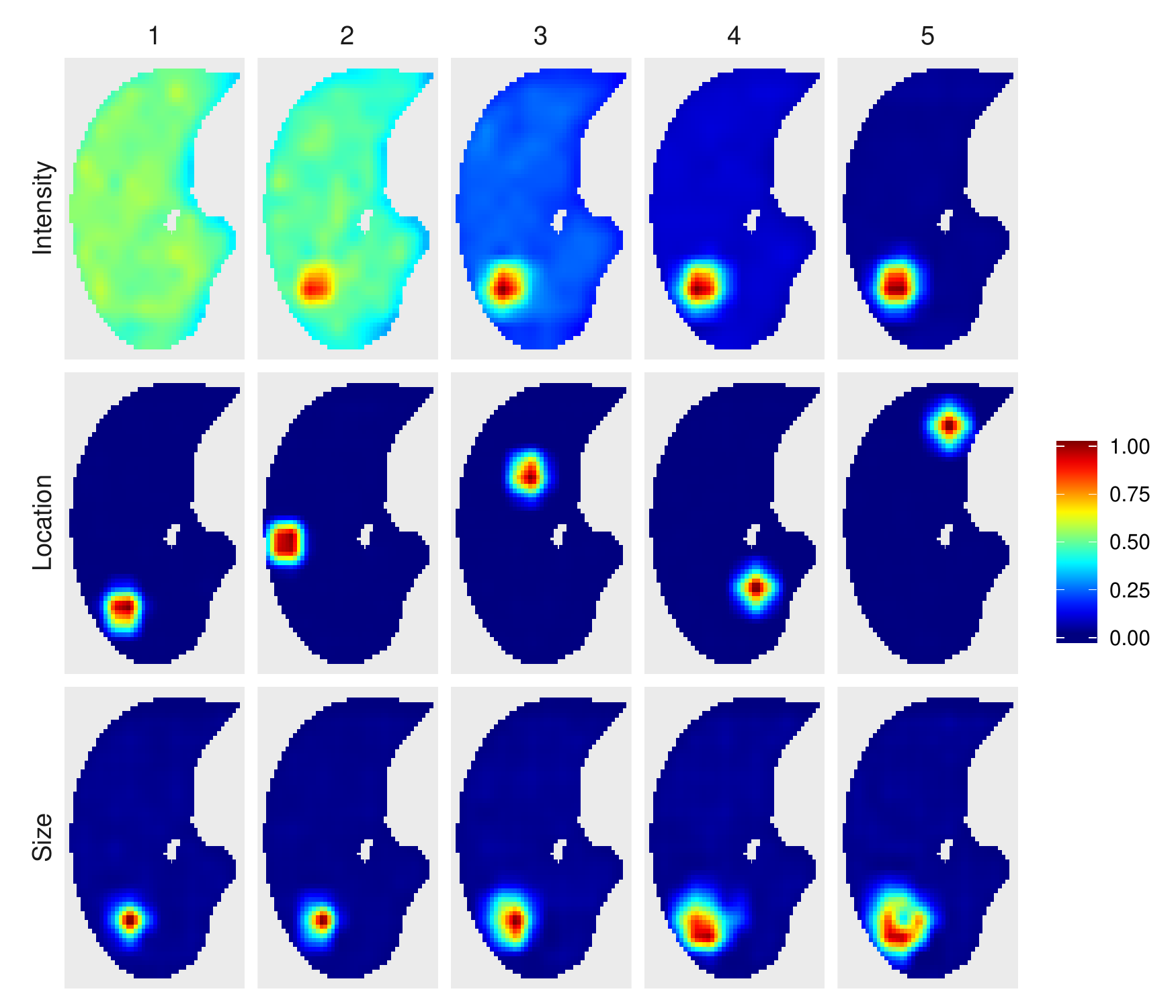}\\
    \caption{Cluster-averaged CLAM (measured in units of importance) for simulation scenarios when the true $K = 5$ and the intensity, location, or size varies across the images. }
    \label{fig:ave_cluster_sim}
\end{figure}

\begin{figure}
    \centering
     \includegraphics[width=0.4\textwidth]{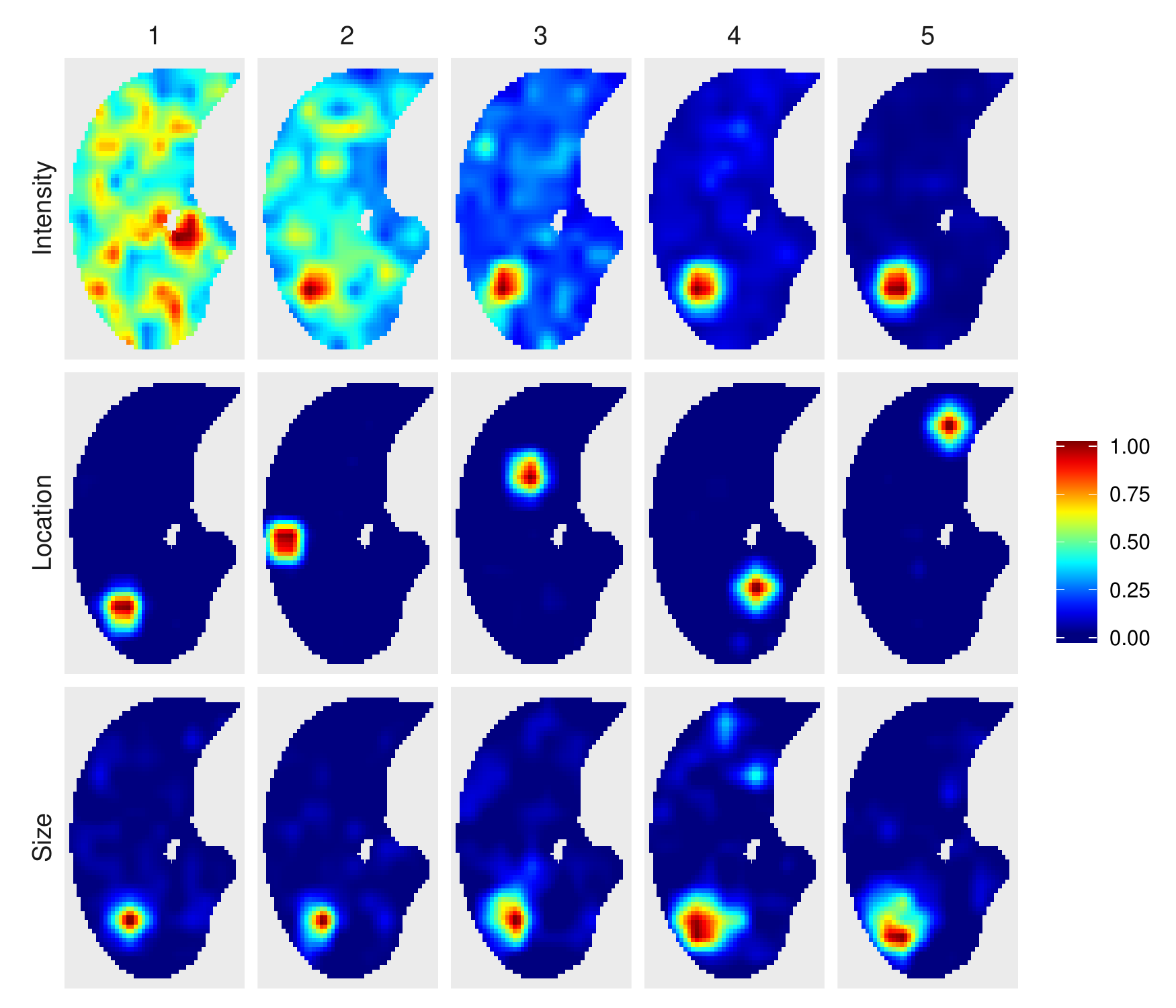} 
    \caption{CLAMs (measured in units of importance) from the simulated lung slices in Figure \ref{fig:design_sim}.}
    \label{fig:ind_clams}
\end{figure}

In Figure \ref{fig:ave_sim}, we show the population-averaged CLAM to obtain a global understanding of discriminative areas for cluster assignments across all images. In Figure \ref{fig:ave_cluster_sim}, we show the cluster-averaged CLAM to identify patterns of discriminative areas within each cluster. Finally, in Figure \ref{fig:ind_clams}, we show the CLAMs for an individual image within each cluster, corresponding to the selected images in Figure \ref{fig:design_sim}. 

For the simulation scenario where intensity was simulated to vary from 0 to 400 HU across five clusters, the population-averaged CLAM (top row of Figure \ref{fig:ave_sim}) indicates that all areas of the lung slice were considered important for cluster assignment to some degree, with a higher average importance in the region with the simulated abnormality compared to the regions where there were no simulated abnormalities, as would be expected. When observing the cluster-averaged CLAMs, different patterns between the clusters emerged. For the images within Cluster 1, which were simulated to have no abnormalities (i.e. 0 HU added to the simulated images), the average CLAM highlighted the entire lung slice as discriminative with an average importance near 0.5 across the lung slice, indicating that different areas of the slice are considered important for different images within the cluster; this was confirmed when observing the CLAM for an individual image as in the top-right subplot of Figure \ref{fig:ind_clams}. In contrast, for the images within Cluster 5, which were simulated to have the highest intensity in the abnormal region, there was only one highlighted region in the cluster-averaged CLAM with the maximum of the average importance at 1 (pixels colored red), indicating that the same area of the slice was considered important for all images within the cluster. This simulation scenario suggests that both the absence and presence of abnormalities within a lung scan can be important for cluster assignments, especially when the intensity is the discriminating characteristic between clusters. Thus, \textit{discriminatory} regions, which are highlighted by CLAM, should not be confused with the actual \textit{abnormal} regions from the lung slice, since healthy (i.e. non-abnormal) regions could be considered discriminatory. 

For the simulation scenario where the location of the abnormality was varied but the size and intensity are constant, the population-averaged CLAM identified five distinct regions of discrimination between the clusters (middle row in Figure \ref{fig:ave_sim}). In the cluster-averaged CLAMs, each cluster was defined by a single discriminatory region at different locations (middle row in Figure \ref{fig:ave_cluster_sim}), indicating that location was the reason for the cluster assignments. Interestingly, although there were no simulated differences in size of abnormality across the images within each cluster, the area of the active discriminatory region from the CLAMs differed significantly across clusters (p$<$0.001); however, the mean HU from the images did not differ across identified clusters (p=0.605), as expected. 

For the simulation scenario where the size of the abnormality was varied from 2 to 6 pixel radii across five clusters, the population-averaged CLAM identified the single region of interest important for cluster assignment (bottom row in Figure \ref{fig:ave_sim}). In the cluster-averaged CLAMs, the same region was also highlighted on each clusters' activation maps, indicating that location was not the reason for the cluster assignments. To supplement the CLAMs, we found that the activated area increased significantly in size (p$<$0.001) from cluster 1 (smallest simulated abnormality) to cluster 5 (largest simulated abnormality), as expected. This simulation scenario suggests that when the cluster-averaged CLAMs are similar across all clusters, although the exact location of discrimination is identified, additional information beyond the CLAM, such as the mean HU or area of activation, can be helpful to understand the reason for the cluster discrimination; in this case, it was the size of the abnormality, but in other cases, it may be the texture or the intensity.

\section{Application}
\label{camapp}
\subsection{Data Collection and Processing}

The sarcoidosis population used in this study was recruited as part of the NHLBI funded Genomic Research in Alpha-1 Antitrypsin Deficiency and Sarcoidosis (GRADS) study. The GRADS study is a multi-center, observational cohort exploring the role of the microbiome and genome in subjects with alpha-1 antitrypsin deficiency and/or sarcoidosis \cite{moller2015rationale}. Subjects were eligible for GRADS if they were between the ages of 18 and 85 years, had a confirmed diagnosis of sarcoidosis via biopsy or manifestations consistent with acute sarcoidosis (Lofgren's syndrome), met one of the nine study phenotypes and provided signed informed consent \cite{moller2015rationale}. As part of GRADS, uniform clinical data was obtained including pulmonary function testing, a chest radiograph (for Scadding staging classification), and a research chest high resolution CT based on the COPDGene protocol \cite{regan2011genetic}. Research chest HRCTs used in this study were acquired under a variety of scanner manufacturers, with the participants in the supine position during breath holding at end inspiration, and the following parameters: 500 msec exposure time, standard B35f kernel, approximately 0.75 mm thickness, and computed interval of approximately 0.5 mm \cite{zach2012quantitative}. 

\begin{table}
\centering
\caption{Demographic features from sarcoidosis population}
\scriptsize
\label{tab:demog}
\begin{tabular}{lrrrr}
  \hline
 & Overall & Cluster 1 & Cluster 2 & P-value \\ 
  \hline
Sample Size &   301 &   126 &   175 &  \\ 
  Male (\%) &   142 (47.2)  &    51 (40.5)  &    91 (52.0)  &  0.063 \\ 
  White (\%) &   218 (72.4)  &    77 (61.1)  &   141 (80.6)  & $<$0.001 \\ 
  Hispanic (\%) &    14 ( 4.7)  &     8 ( 6.4)  &     6 ( 3.4)  &  0.355 \\ 
  Age (mean (sd)) & 52.87 (9.75) & 53.58 (9.32) & 52.36 (10.04) &  0.287 \\ 
  BMI (mean (sd)) & 30.56 (6.55) & 31.05 (7.05) & 30.20 (6.16) &  0.267 \\ 
  Height (mean (sd)) & 67.08 (4.15) & 66.52 (3.96) & 67.49 (4.24) &  0.045 \\ 
  Model (\%) &    &     &    &  0.929 \\ 
      \phantom{000}Discovery CT750 &     4 ( 1.3)  &     1 ( 0.8)  &     3 ( 1.7)  &  \\ 
     \phantom{000}iCT 128 &    26 ( 8.6)  &    10 ( 7.9)  &    16 ( 9.1)  &  \\ 
     \phantom{000}LightSpeed VCT &   101 (33.6)  &    45 (35.7)  &    56 (32.0)  &  \\ 
      \phantom{000}Optima CT660 &     3 ( 1.0)  &     2 ( 1.6)  &     1 ( 0.6)  &  \\ 
     \phantom{000}Sensation 64 &     8 ( 2.7)  &     4 ( 3.2)  &     4 ( 2.3)  &  \\ 
     \phantom{000}Definition &    38 (12.6)  &    16 (12.7)  &    22 (12.6)  &  \\ 
    \phantom{000}Definition AS+ &    87 (28.9)  &    36 (28.6)  &    51 (29.1)  &  \\ 
     \phantom{000}Definition Flash &    34 (11.3)  &    12 ( 9.5)  &    22 (12.6)  &  \\ 
  Scadding (\%) &    &     &    & $<$0.001 \\ 
      \phantom{000}0 &    40 (13.3)  &     7 ( 5.6)  &    33 (18.9)  &  \\ 
      \phantom{000}1 &    59 (19.6)  &    17 (13.5)  &    42 (24.0)  &  \\ 
      \phantom{000}2 &    87 (28.9)  &    32 (25.4)  &    55 (31.4)  &  \\ 
      \phantom{000}3 &    43 (14.3)  &    16 (12.7)  &    27 (15.4)  &  \\ 
      \phantom{000}4 &    72 (23.9)  &    54 (42.9)  &    18 (10.3)  &  \\ 
   \hline
\end{tabular}
\end{table}

Images were obtained in raw DICOM (Digital Imaging and Communications in Medicine) format, and converted to three-dimensional NIfTI (Neuroimaging Informatics Technology Initiative) using dcm2niix (\url{https://github.com/rordenlab/dcm2niix}) from the dcm2niir R package. We resampled all scans to 1 $mm^3$ format, to normalize scans to the same resolution. Next, we segmented the left and right lungs using segment\_lung\_lr from the lungct R package (\url{https://github.com/ryansar/lungct}).  To create a study-specific template, we used a random sample of 10\% of the GRADS subjects, then followed the template creation protocol from \cite{ryan2019template}; the template converged in 9 iterations. Finally, we performed SyN non-linear registration to the study-specific template via antsRegistration from the ANTsR R package to align all scans to the same space and orientation, removing positional inconsistencies between subjects. To address memory limitations in computation, we further resampled the scans down to 3$mm^3$ voxel spacing, the same spacing as in our simulations.

\subsection{Statistical Methodology}

We used the entire 3D lung scan with both the right and left lungs to identify clusters within our images. We used a similar CAE as in our simulations, but changed the 2D convolutional layers to 3D and reduced the second convolutional layer from a stride length of 1 to 2, to reduce the number of learned parameters which can become very large in the 3D space; then, the 3D input image was of dimension $(88\times88\times88)$, with the output of the last convolutional layer at $(11\times11\times11)$. We estimated the optimal number of clusters prior to the clustering layer of our framework using the Silhouette method \cite{rousseeuw1987silhouettes} with sparse k-means \cite{witten2010framework}. Then, setting the number of clusters to the optimal estimate, we identified clusters within these data, using CLAM to localize the discriminative regions. To evaluate the usefulness of the new subtypes identified with our clustering methodology, we found the linear associations with our clusters and various clinical outcomes. Clinical outcomes included lung function using the forced expiratory volume at one second (FEV1), forced vital capacity (FVC), diffusing capacity for carbon monoxide (DLCO), laboratory testing including CD4 cell count and C-Reactive protein (CRP), and patient reported outcomes, including fatigue assessment scale (FAS), gastroesophageal reflux disease questionnaire (GERDQ), cognitive failure questionnaire (CFQ), shortness of breath questionnaire (SOBQ), Promis, and SF-12 questionnaire.

\subsection{Application Results}

In our study population, we had n=301 subjects with sarcoidosis, of which 47.2\% were male, 72.9\% were white, and 4.7\% were Hispanic. The average age was 53 years (SD=9.75 years), with an average height of 67 inches (SD=4.15 inches) and an average BMI of 30.58 (SD=6.55) (Table \ref{tab:demog}). The optimal number of clusters within the sarcoidosis population was estimated at two, with 126 subjects in Cluster 1 and 175 subjects in Cluster 2. Subjects assigned to each cluster had a similar age (p=0.287) and BMI (p=0.267), and no significant differences in sex (p=0.063), ethnicity (p=0.355), nor scanner models (p=0.929). Subjects in Cluster 1 had a lower proportion of whites (61\% vs 81\%) compared to that of Cluster 2. There was also a significant association between the new clusters and the existing Scadding stage classification (p$<$0.001), with more subjects in Cluster 1 classified as Scadding stage 4 (43\% vs. 10\%) and more of those in Cluster 2 with Scadding stage 0 and 1.  

\begin{figure}
    \centering
    (a)
    \includegraphics[width=0.2\textwidth]{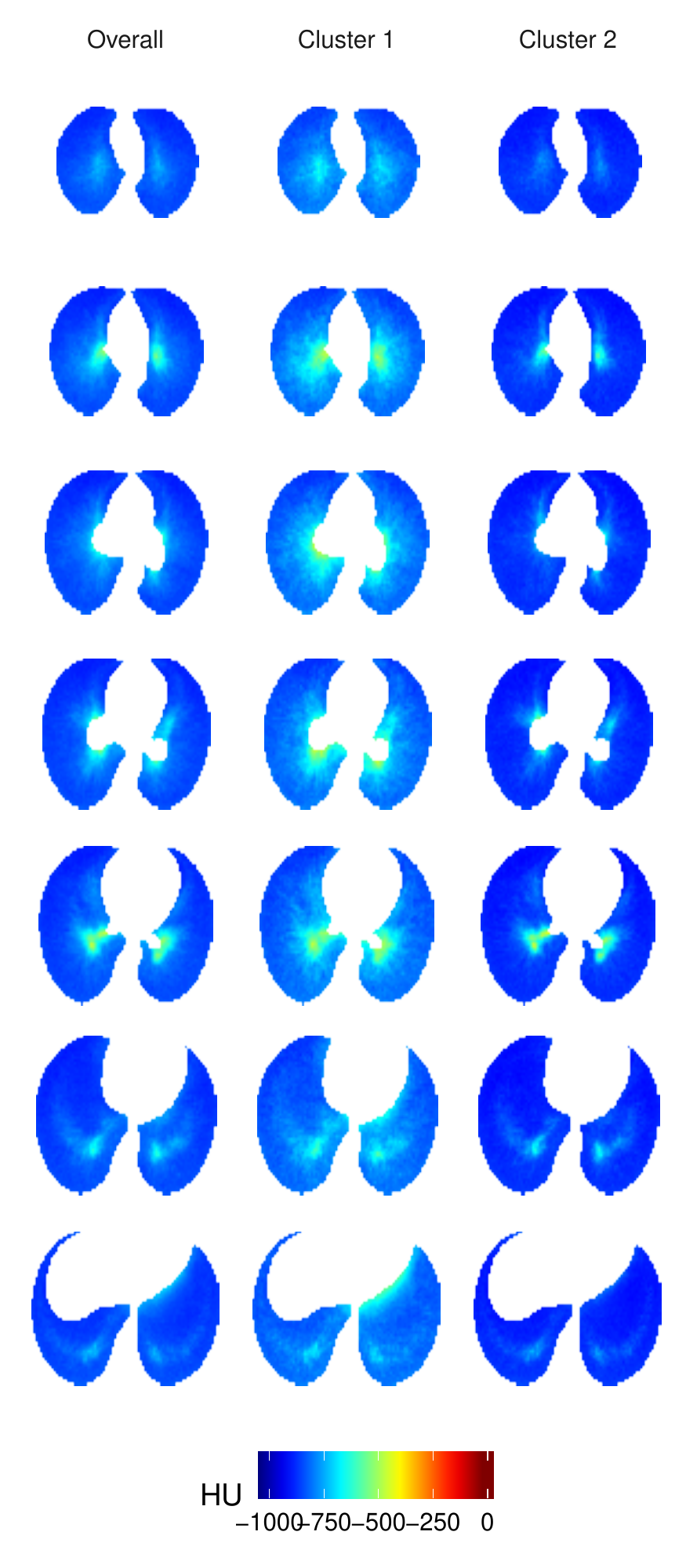}
    (b)
    \includegraphics[width=0.2\textwidth]{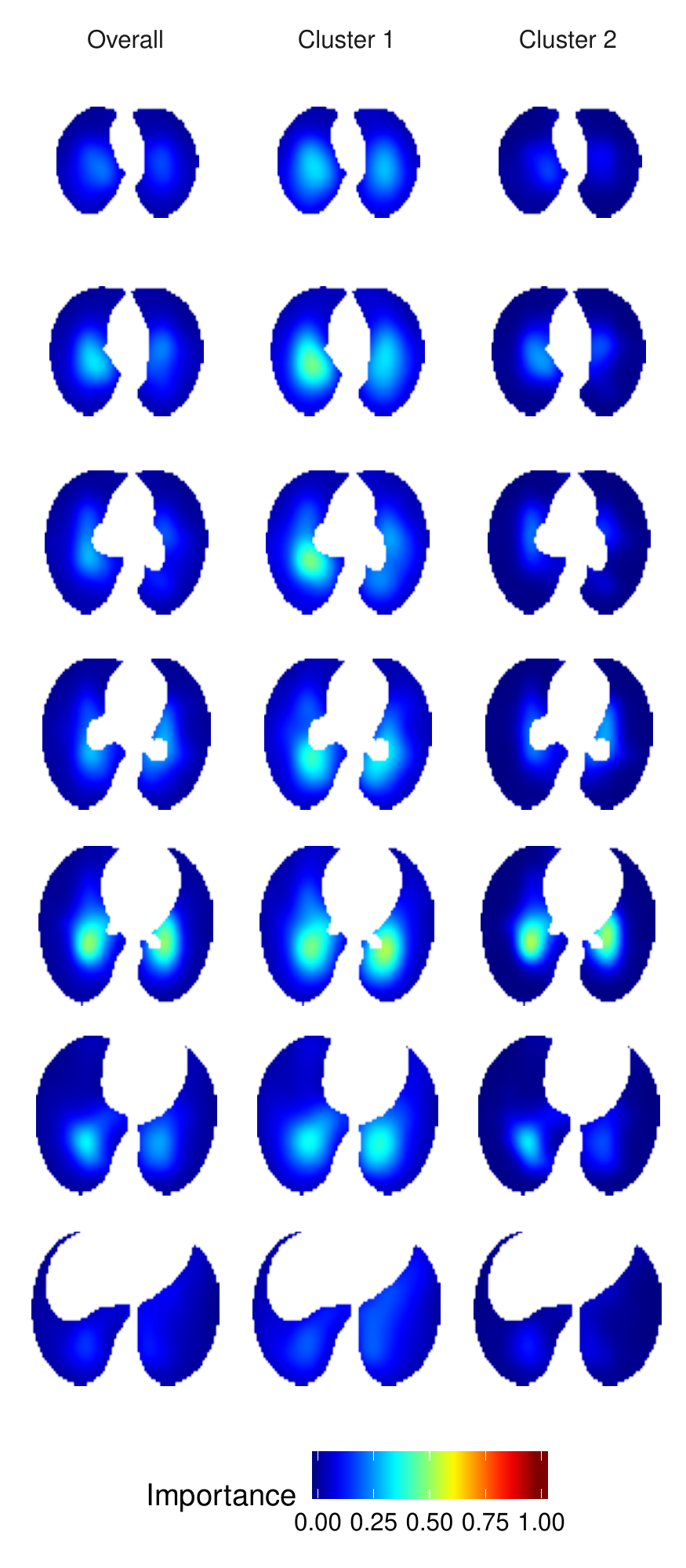}
    \caption{(a) Average CT scans from all patients and split by each cluster. (b) Population- and cluster-averaged CLAMs. Various 2D axial slices of the 3D image are shown from the top of the lung to the bottom. The 2D slices are in radiological convention, with the right lung on the left side and the front of the lung at the top.}
    \label{fig:ave_3D}
\end{figure}

The average CT scans across all patients are shown in Figure \ref{fig:ave_3D} (a), and the average CLAMs are shown in Figure \ref{fig:ave_3D} (b). As seen by the average scan across the entire study population, there were higher HUs near the hilar region, where the airways and blood vessels enter the lungs and lymph nodes are often enlarged. In the average scan for Cluster 1, we saw larger areas of increased HU extending beyond the hilar region, particularly in the upper slices of the lung, which was not observed in the average scan for Cluster 2 (Figure \ref{fig:ave_3D} (a)). Further, in the average CLAM for Cluster 1, we saw areas of discrimination throughout the lung, with stronger importance in the upper lobes of the lung, whereas the average areas of discrimination for images assigned to Cluster 2 appeared to be more localized to the hilar region (Figure \ref{fig:ave_3D} (b)). Additionally, throughout much of the lung, the average CLAM for Cluster 2 had an importance of 0 (pixels colored dark blue), indicating that there were large portions of the CT scans from the patients in Cluster 2 that have no discriminating characteristics. These cluster-averaged CLAMs suggest that images in Cluster 1 had more discriminative areas throughout the lung, with common areas of discrimination appearing in the upper lung as well as near the hilar region, as compared with the images in Cluster 2 which had less discriminative areas throughout the lung, with the only common area of discrimination in the hilar region. 

\begin{figure}
    \centering
    \includegraphics[width=0.4\textwidth]{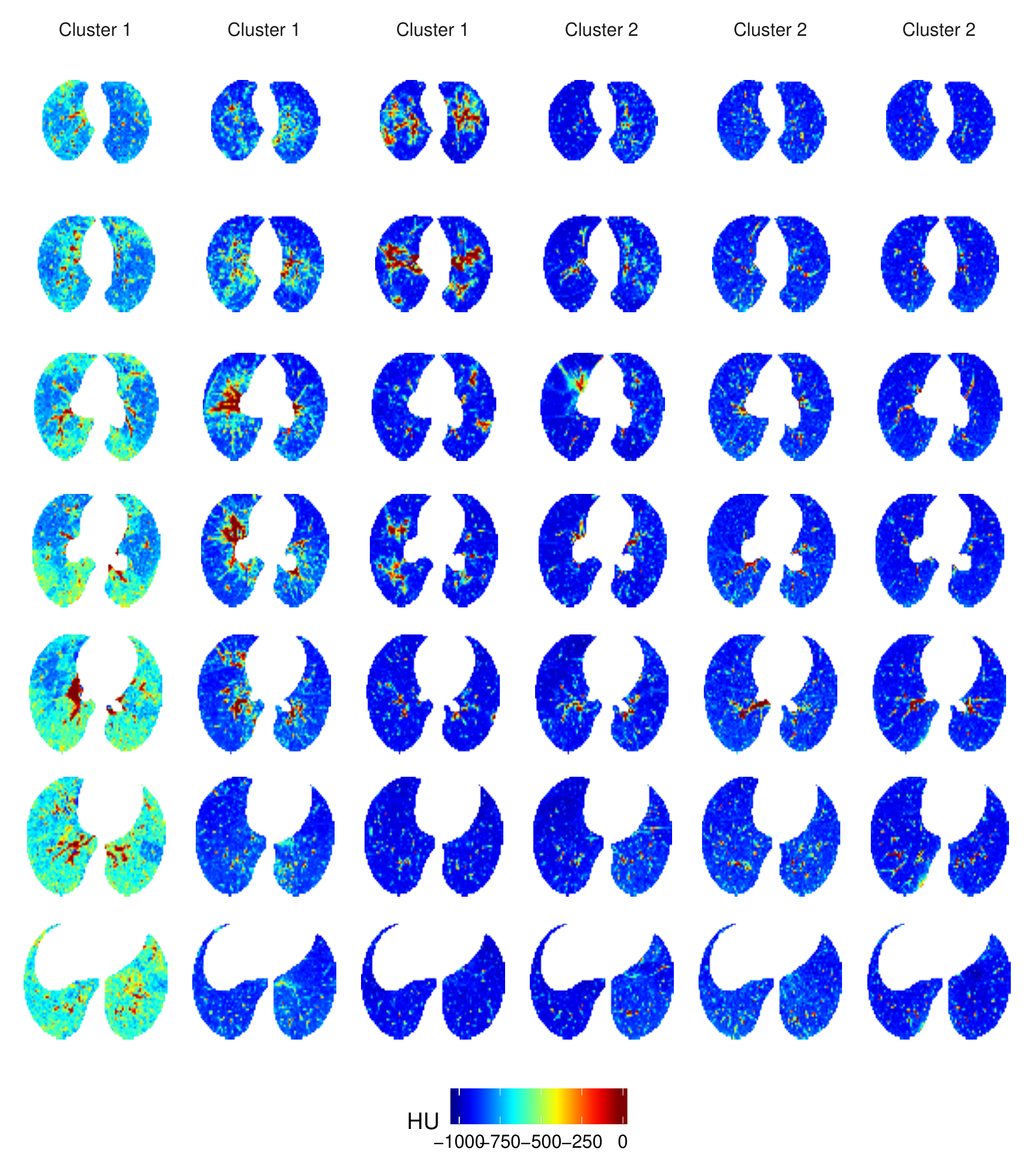}\\
    \caption{CT scans from three randomly selected subjects from each cluster. Various 2D axial slices of the 3D image are shown from the top of the lung to the bottom. The 2D slices are in radiological convention, with the right lung on the left side and the front of the lung at the top.}
    \label{fig:ind_3D}
\end{figure}

\begin{figure}
    \centering
    \includegraphics[width=0.4\textwidth]{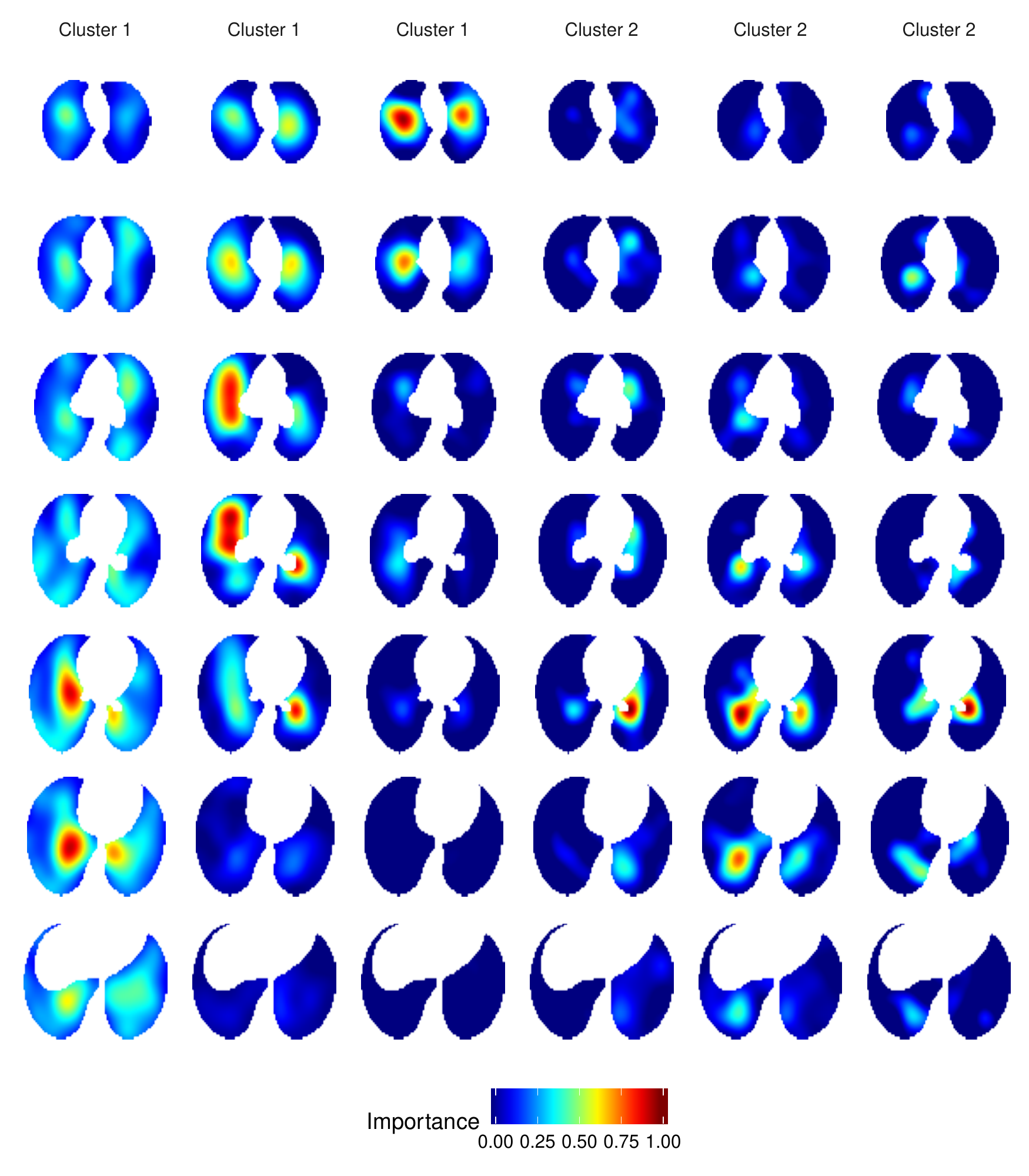}\\
    \caption{CLAMs from three randomly selected subjects from each cluster. Various 2D axial slices of 3D image are shhown from the top of the lung to the bottom. The 2D slices are in radiological convention, with the right lung on the left side and the front of the lung at the top.}
    \label{fig:ind_cluster_3D}
\end{figure}

To better understand the cluster assignments for each individual CT scan, we looked at randomly selected scans and CLAMs from each cluster (Figures \ref{fig:ind_3D} and \ref{fig:ind_cluster_3D}). As is observed from the first selected CT scan from Cluster 1 (left column of Figure \ref{fig:ind_3D}), much of the lung shows increased HU (i.e. green areas), especially in the lower slices of the lung, with the highest HU in the right lung (i.e. red areas). In the corresponding CLAM (left column of Figure \ref{fig:ind_cluster_3D}), the lower lung slices on the right lung had the highest importance for cluster assignment, with all other areas having smaller, yet non-zero, importance values, indicating these regions were also important for cluster assignment. Furthermore, while the other two randomly selected scans from Cluster 1 (second and third columns of Figure \ref{fig:ind_3D}) had less abnormalities throughout the lung, they both showed certain areas with abnormalities; for the second scan, the abnormal areas were near the center of the lung, whereas, for the third scan, the abnormal areas were near the top of the lung. For all three of the selected scans in Cluster 1, the corresponding CLAMs highlighted the areas of increased HU, where increased HU is associated with advanced abnormalities in pulmonary sarcoidosis. 

For the randomly selected scans in Cluster 2, most of the voxels were within a healthy lung range (approximately -900 to -700 HU, roughly) (Figures \ref{fig:ind_3D}). There were certain spots on the scans with increased HU, which could be a combination of the airways and vasculature which are common to all lung scans, or they could suggest abnormalities as before. However, on the corresponding CLAMs (Figure \ref{fig:ind_cluster_3D}), the areas near the hilar region were most important for cluster assignment, which indicates that either the presence of absence of abnormalities in this location influence the cluster assignment, which differed from the images in Cluster 1 where more regions were influential. 

\begin{figure}
    \centering
    \includegraphics[width=0.4\textwidth]{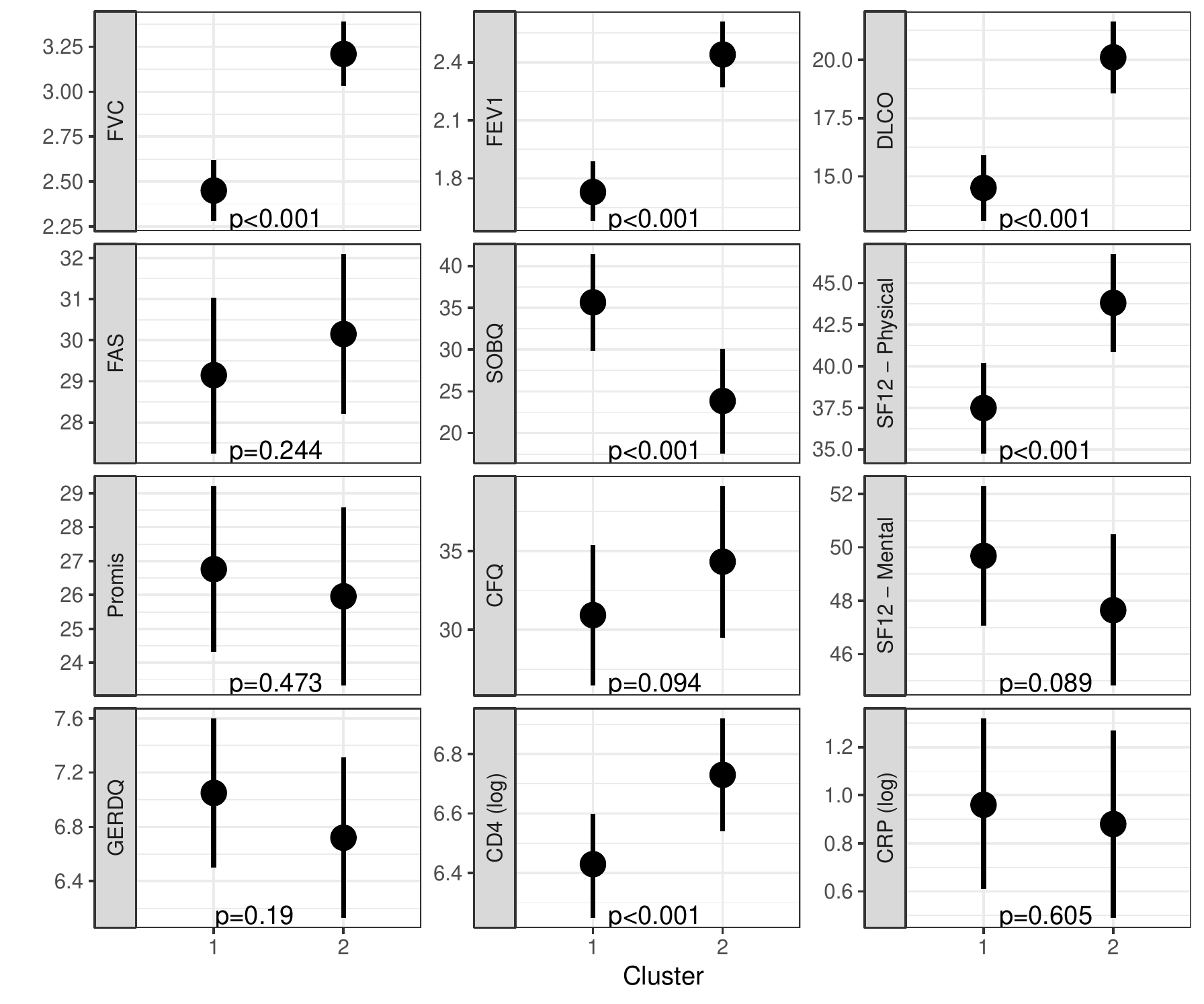}\\
    \caption{Associations with pulmonary function tests and patient reported outcomes }
    \label{fig:pros_3d}
\end{figure}

To determine whether the newly identified clusters had any clinical value, we assessed the linear association with pulmonary function tests and patient reported outcomes. We found that the new clusters showed significant associations with all pulmonary function tests (p-values for FVC, FEV1, DLCO $<$0.001) and some patient reported outcomes, including SOBQ, SF12-Physical, and CD4 count (p$<$0.001) (Figure \ref{fig:pros_3d}). Further, the addition of the new clustering variable explained more variation for all outcomes, except for CRP and PROMIS, compared to the base model with only demographics. Additionally, the new clustering variable, based on the learned features within the 3D CT scans, explained more variation than Scadding for FEV1, FVC, DLCO, FAS, GERDQ, CFQ, SOBQ, PROMIS, and SF12-Physical \& Mental outcomes.

\section{Conclusion}
\label{camconc}
In this work, we developed cluster activation mapping by modifying the architecture of the neural network and the weighting algorithm from Score-CAM, which was originally developed for classification. Our method identifies cluster-specific discriminative regions from images, representing a novel contribution in understanding cluster assignments from deep clustering architectures. The performance of our approach has been evaluated using a simulation study and applied to 3D CT scans from a sarcoidosis population to identify new clusters of sarcoidosis based purely on CT scan presentation, which could lead to new subtypes of this disease. 

In our simulation study, we found that clusters based on location of abnormality on the images are easily identifiable using CLAM. Additionally, we found that both the presence and absence of abnormalities can be discriminatory for cluster assignment, as seen in our simulations when a cluster of scans that had no added abnormality had CLAMs highlighting the entire lung region as discriminatory. Finally, for the simulations based on size of abnormality, we showed how CLAMs can appear similar across all clusters even though a true difference between the clusters exists. In these cases, CLAM will still identify the discriminatory region even if it is in the same location across the clusters. However, to find the reason behind the differences in clusters, additional characteristics beyond the CLAMs from the original images are needed.

In our application to a sarcoidosis population, we identified two clusters of subjects based on their CT scans. Our CLAM technique identified that the scans assigned to the clusters differed in their amount and location of discriminative regions across the lung slice. Subjects in the cluster identified by the cluster-averaged CLAMs as having more discriminatory regions throughout the lung, particularly in the upper lung slices as noted in Cluster 1, were shown to have significantly reduced lung function and scored significantly worse on their shortness of breath and SF-12 physical questionnaires compared to the subjects assigned to the other cluster (Cluster 2) with less discriminatory regions throughout the lung. This is likely due to more significant abnormalities spread across the lung in the former group, impacting lung functionality and resulting in poor outcomes, both from a lung function and patient reported outcome. Furthermore, these newly identified clusters explained more variation in the majority of the clinical outcomes than Scadding stage, an existing classification of sarcoidosis subjects based on the visual assessment of chest radiography. The finding of reduced CD4 count in this group is also consistent with more significant disease in prior studies \cite{sweiss2010significant}. Thus, the clusters obtained under the deep clustering framework and visualized using CLAM are clinically meaningful, even more so than the existing clusters based on Scadding stage. With additional research, these new clusters could be developed into a novel disease subtype for sarcoidosis.

The major contribution of CLAM is that it increases the transparency behind an image's cluster assignment in a way that humans can interpret; that is, it produces heatmaps of discriminatory regions. As deep learning networks are often criticized and not trusted due to their lack of interpretability, this contribution to deep clustering architectures should not be underestimated. Existing methods to cluster medical imaging data use visual scoring assessments for this reason, that they are interpretable to the human eye. Yet, we know that visual assessments can (1) be subjective and biased from the reader, (2) be costly and slow to obtain, and (3) miss patterns within the data that are not visually detectable. Thus, this work helps to make clustering assignments from deep clustering architectures more transparent, providing a data-driven method for image clustering, which can ultimately lead to new subtypes of various diseases using medical imaging datasets.

An existing data-driven approach for clustering of images, different from our deep clustering approach, is based on the pre-specified, classical radiomic features, which consists of hundreds of highly-correlated quantitative features extracted from the images \cite{afshar2018hand}. While they have proven useful in certain applications \cite{halder2020lung}, the large dimension of highly-correlated features common to these radiomic analyses presents challenges. Furthermore, clustering based on these features are not able to detect differences in images based on location of the abnormality, and does not provide a visual understanding of why images are assigned to a particular cluster. Thus, our ``learned" features which we extract here in our deep clustering framework, can be advantageous since (1) the features are learned from the data, providing the best representation of the data in a reduced dimension, (2) we can detect differences between clusters based on a variety of characteristics, including intensity, location and size, among others, and (3) with the addition of CLAM, we are able to provide a visual tool to aid in the understanding of the cluster assignments.

Besides the development of CLAM for a single image, another contribution of this work is in the group-averaged CLAMs. For many applications in image processing, a group averaged activation map is nonsensical, since images may not be aligned prior to deep clustering. Even if images were aligned in some way, common locations of activation across the images may not be important (e.g. if trying to cluster dogs and cats, the location of the animal in the image is not critical). However, in medical imaging, scans are registered to a common template \cite{ryan2019template} to align the anatomy across the images. Often, disease can manifest itself in common locations across the scan \cite{worsley1996unified}. Thus, in the case of medical imaging data, group-averaged CLAMs have clinical value, and give us a population- or cluster-level understanding of the discriminative regions from the cluster assignments.

For estimating the optimal number of clusters, we concluded that this number should be estimated prior to the clustering layer, due to changes in the feature space caused by the clustering layer. However, a limitation of this work is the lack of a clear method to estimate the optimal number of clusters from the data, which is true for most clustering algorithms. In our simulation section, we used an existing approach, the average silhouette method, which correctly estimated the number of clusters the majority of the time. As part of a sensitivity analysis, we also explored the Gap statistic and found that the Silhouette had better performance under simulation. However, more research on selecting the optimal number of clusters is needed. 

Another limitation of this work is that the CLAMs are only as good as the clustering algorithm. That is, if the deep clustering does a poor job of finding distinct clusters, the CLAMs will not necessarily identify the correct discriminatory regions. Techniques to improve the accuracies of deep clustering architectures may include data augmentation \cite{guo2018deep}, generative adversarial networks \cite{berthelot2018understanding} and alternative loss functions \cite{mrabah2019deep}, among others. 

In conclusion, we developed cluster activation mapping to identify cluster-specific discriminative regions from images, which has been evaluated under simulation and applied to 3D CT scans from a diseased population of sarcoidosis subjects. Our approach represents a novel contribution in the understanding of cluster assignments from deep clustering architectures, and provides an alternative and superior data-driven approach for clustering of medical imaging data beyond visual assessment or the classical radiomic approaches. The clusters identified and understood with our methodology can ultimately lead to the identification of new subtypes for various diseases.

This work was supported by the National Institutes of Health (R01 HL114587; R01 HL142049; U01 HL112695). Data from the GRADS study was used, which was funded by the NIH grant U01 HL112707 entitled ``Sarcoidosis and A1AT Genomics and Informatics Center'', as well as others (U01 HL112707, U01 HL112694, U01 HL112695, U01 HL112696, U01 HL112702, U01 HL112708, U01 HL112711, U01 HL112712). The content is solely the responsibility of the authors and does not necessarily represent the official views of the National Heart, Lung, and Blood Institute or the National Institutes of Health.

\bibliographystyle{ieeetr}
\bibliography{ref.bib}

\end{document}